\documentclass[11pt,letterpaper]{article}
\usepackage{naaclhlt2016}
\usepackage{pslatex}
\usepackage{latexsym}

\usepackage{multirow}
\usepackage{tikz}
\usetikzlibrary{arrows,shapes,trees}
\usepackage{fixltx2e}
\usepackage{amsmath}
\usepackage{amssymb}
\usepackage{enumitem}
\usepackage{bold-extra}
\usepackage{rotating,url}
\usepackage{tikz-dependency}

\naaclfinalcopy 
\def\naaclpaperid{164} 

\title{Top-down Tree Long Short-Term Memory Networks}

\author{Xingxing Zhang, Liang Lu \and Mirella Lapata \\
         School of Informatics, University of Edinburgh \\  10 Crichton Street, Edinburgh EH8 9AB, UK \\
         {\tt \{x.zhang,liang.lu\}@ed.ac.uk,mlap@inf.ed.ac.uk} }

\date{}

\begin{document}

\maketitle

\begin{abstract}
  Long Short-Term Memory (LSTM) networks, a type of recurrent neural
  network with a more complex computational unit, have been
  successfully applied to a variety of sequence modeling tasks.  In
  this paper we develop Tree Long Short-Term Memory (\mbox{\sc
    TreeLSTM}), a neural network model based on LSTM, which is
  designed to predict a tree rather than a linear sequence. 
  {\sc TreeLSTM} defines the probability of a sentence by estimating
  the generation probability of its dependency tree. At each time
  step, a node is generated based on the representation of the
  generated sub-tree. We further enhance the modeling power of
  \mbox{\sc TreeLSTM} by explicitly representing the correlations
  between left and right dependents.  Application of our model to the
  MSR sentence completion challenge achieves results beyond the
  current state of the art. We also report results on dependency
  parsing reranking achieving competitive performance.

\end{abstract}

\section{Introduction}


Neural language models have been gaining increasing attention as a
competitive alternative to \mbox{n-grams}.  The main idea is to
represent each word using a real-valued feature vector capturing the
contexts in which it occurs. The conditional probability of the next
word is then modeled as a smooth function of the feature vectors of
the preceding words and the next word. In essence, similar
representations are learned for words found in similar contexts
resulting in similar predictions for the next word.  Previous
approaches have mainly employed feed-forward
\cite{bengio2003neural,MnihHinton2007} and recurrent neural networks
\cite{mikolov2010recurrent,mikolov2012thesis} in order to map the
feature vectors of the context words to the distribution for the next
word. Recently, RNNs with Long Short-Term Memory (LSTM) units
\cite{hochreiter1997long,Hochreiter:1998} have emerged as a popular architecture due to
their strong ability to capture long-term dependencies.  LSTMs have
been successfully applied to a variety of tasks ranging from machine
translation \cite{sutskever2014sequence}, to speech recognition
\cite{graves2013speech}, and image description generation
\cite{Vinyals_2015_CVPR}.

Despite superior performance in many applications, neural language
models essentially predict sequences of words.  Many NLP tasks,
however, exploit syntactic information operating over tree structures
(e.g.,~dependency or constituent trees). In this paper we develop a
novel neural network model which combines the advantages of the LSTM
architecture and syntactic structure.  Our model estimates the
probability of a sentence by estimating the generation probability of
its dependency tree. Instead of explicitly encoding tree structure as
a set of features, we use four LSTM networks to model four types of
dependency edges which altogether specify how the tree is built. At
each time step, one LSTM is activated which predicts the next word
conditioned on the sub-tree generated so far.  To learn the
representations of the conditioned sub-tree, we force the four LSTMs
to share their hidden layers.
Our model is also capable of generating trees just by sampling from a
trained model and can be seamlessly integrated with text generation
applications.

Our approach is related to but ultimately different from recursive
neural networks \cite{Pollack:1990} a class of models which operate on
structured inputs.  Given a (binary) parse tree, they recursively
generate parent representations in a \mbox{bottom-up} fashion, by
combining tokens to produce representations for phrases, and
eventually the whole sentence.  The learned representations can be
then used in classification tasks such as sentiment analysis
\cite{socher-EtAl:2011:EMNLP} and paraphrase detection
\cite{Socher:ea:2011}.  \newcite{tai:ea:2015} learn distributed
representations over syntactic trees by generalizing the LSTM
architecture to tree-structured network topologies. The key feature of
our model is not so much that it can learn semantic representations of
phrases or sentences, but its ability to predict tree structure and
estimate its probability.

Syntactic language models have a long history in NLP dating back to
\newcite{Chelba:Jelinek:2000} (see also
\newcite{roark2001probabilistic} and
\newcite{charniak2001immediate}). These models differ in how grammar
structures in a parsing tree are used when predicting the next word.
Other work develops dependency-based language models for specific
applications such as machine translation
\cite{shen2008new,zhang2009structured,sennrich2015modelling}, speech
recognition \cite{chelba1997structure} or sentence completion
\cite{gubbins-vlachos:2013:EMNLP}. All instances of these models apply
Markov assumptions on the dependency tree, and adopt standard n-gram
smoothing methods for reliable parameter
estimation. \newcite{Emami:ea:2003} and
\newcite{sennrich2015modelling} estimate the parameters of a
structured language model using feed-forward neural networks
\cite{bengio2003neural}. \newcite{mirowski-vlachos:2015} re-implement
the model of \newcite{gubbins-vlachos:2013:EMNLP} with RNNs. They view
sentences as sequences of words over a tree. While they ignore the
tree structures themselves, we model them explicitly.

Our model shares with other structured-based language models the
ability to take dependency information into account.  It differs in
the following respects: (a) it does not artificially restrict the
depth of the dependencies it considers and can thus be viewed as an
infinite order dependency language model; (b)~it not only estimates
the probability of a string but is also capable of generating
dependency trees; (c)~finally, contrary to previous dependency-based
language models which encode syntactic information as features, our
model takes tree structure into account more directly via representing
different types of dependency edges explicitly using LSTMs. Therefore,
there is no need to manually determine which dependency tree features
should be used or how large the feature embeddings should be.




We evaluate our model on the MSR sentence completion challenge, a
benchmark language modeling dataset. Our results outperform the best
published results on this dataset. Since our model is
a general tree estimator, we also use it to rerank the top~$K$
dependency trees from the (second order) MSTPasrser and obtain
performance on par with recently proposed dependency parsers. 



\section{Tree Long Short-Term Memory Networks}

We seek to estimate the probability of a sentence by estimating the
generation probability of its dependency tree. Syntactic information
in our model is represented in the form of dependency
paths. 
In the following, we first describe our definition of dependency path
and based on it explain how the probability of a sentence is
estimated.

\subsection{Dependency Path}
\label{sec:dpath}

\begin{figure}[t]
\centering
\begin{tikzpicture}[scale=.5,->,>=stealth',thick,main node/.style={circle,fill=blue!20,draw,inner sep=0pt,minimum size=3mm}]
\node[main node][label=right:$w_0$] (L0) at (-1, 3) {};
\node[main node][label={[xshift=-2pt]$w_1$}] (L1) at (-3, 0) {};
\node[main node][label={[xshift=-4pt]$w_{k-1}$}] (L2) at (-5.5, 0) {};
\node[main node][label={[xshift=3pt]$w_k$}] (L3) at (-9.5, 0) {};
\node[main node][label=$w_n$] (L4) at (-12, 0) {};

\path (L0) edge node[below right= 1pt] {\textbf{\textsc{Left}}} (L1);
\path (L1) edge[-,line width=1.2pt,style=dotted] (L2);
\path (L2) edge node[below = 5pt] {\textbf{ \textsc{ Nx-Left } }} (L3);
\path (L0) edge[bend right = 17, style=dashed]  (L2);
\path (L0) edge[bend right = 13, style=dashed]  (L3);
\path (L0) edge[bend right = 11, style=dashed]  (L4);
\path (L3) edge[-,line width=1.2pt,style=dotted] (L4);
\end{tikzpicture}
\vspace{-3.5mm}
\caption{{\sc Left} and {\sc Nx-Left} edges. Dotted line between $w_1$
  and $w_{k-1}$ (also between~$w_k$ and~$w_n$) indicate that there may
  be~\mbox{$\geq 0$} nodes inbetween.}
\label{fig:etype}
\end{figure}
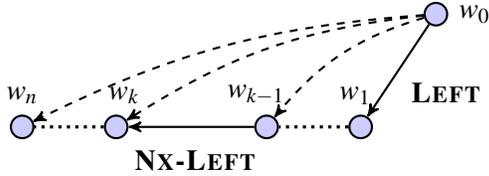

Generally speaking, a dependency path is the path between {\sc root}
and~$w$ consisting of the nodes on the path and the edges connecting
them.
To represent dependency paths, we introduce four types of edges which
essentially define the ``shape'' of a dependency tree. Let~$w_0$
denote a node in a tree and~$w_1, w_2, \dots, w_n$ its left
dependents. As shown in Figure~\ref{fig:etype}, {\sc Left} edge is the
edge between $w_0$~and its first left dependent denoted as~$(w_0,
w_1)$.  Let~$w_k$ (with \mbox{$1 < k \leq n$}) denote a non-first left
dependent of~$w_0$. The edge from $w_{k-1}$ to $w_k$ is a \mbox{{\sc
    Nx-Left}} edge ({\sc Nx} stands for {\sc Next}), where~$w_{k-1}$ is the right adjacent sibling of
$w_k$. Note that the {\sc Nx-Left} edge $(w_{k-1}, w_k)$ replaces edge
$(w_0, w_k)$ (illustrated with a dashed line in
Figure~\ref{fig:etype}) in the original dependency tree. The
modification allows information to flow from $w_0$ to $w_k$
through~\mbox{$w_1, \dots, w_{k-1}$} rather than directly from~$w_0$
to $w_k$. {\sc Right} and {\sc Nx-Right} edges are defined analogously
for right dependents.

Given these four types of edges, dependency paths (denoted as $\mathcal{D}(w)$) can be defined as
follows bearing in mind that the first right dependent of {\sc root}
is its only dependent and that~$w^p$ denotes the parent of~$w$. We use $( \dots )$ to denote a sequence, where $()$ is an empty sequence and $\Vert$ is an operator for concatenating two sequences.
\begin{enumerate}[label={(\arabic*)}]
\itemsep-.3em
\item if $w$ is {\sc root}, then $\mathcal{D}(w) = ()$ 
\item if $w$ is a {\bf left dependent} of $w^p$
\vspace*{-.2cm}
\begin{enumerate}
\itemsep-.1em
 \item  if $w$ is the first left dependent, then 
$\mathcal{D}(w) = \mathcal{D}(w^p) \Vert ( \langle w^p, \text{\sc Left}\rangle )$ 
  \item  if $w$ is not the first left dependent and
$w^s$ is  its right adjacent sibling, then \\
$\mathcal{D}(w) = \mathcal{D}(w^{s}) \Vert ( \langle w^s,
\text{\sc Nx-Left}\rangle )$
\end{enumerate}
\vspace*{-.1cm}
\item if $w$ is a {\bf right dependent} of $w^p$ 
\vspace*{-.2cm}
\begin{enumerate}
\item if $w$ is the first right dependent, then 
$\mathcal{D}(w) = \mathcal{D}(w^p) \Vert ( \langle w^p, \text{\sc Right}\rangle
)$ 
\item  if $w$ is not the first right dependent and
$w^s$ is its left adjacent sibling, then \\  
$\mathcal{D}(w) = \mathcal{D}(w^s) \Vert ( \langle w^s, \text{\sc
    Nx-Right}\rangle )$ 
\end{enumerate}
\end{enumerate}
%
A dependency tree can be represented by the set of its dependency
paths which in turn can be used to reconstruct the original tree.\footnote{Throughout this paper we assume all dependency trees are projective.}

Dependency paths for the first two levels of the tree in Figure~\ref{fig:deptree} are as follows
(ignoring for the moment the subscripts which we explain in the next
section). $\mathcal{D}(\text{sold}) = ( \langle\text{\sc root},
\text{\sc Right}\rangle )$ (see definitions~(1) and~(3a)),
$\mathcal{D}(\text{year})=\mathcal{D}(\text{sold}) \Vert
( \langle\text{sold}, \text{\sc Left}\rangle )$~(see~(2a)), 
$\mathcal{D}(\text{manufacturer}) = \mathcal{D}(\text{year}) \Vert
( \langle\text{year}, \text{\sc Nx-Left}\rangle )$ (see~(2b)),
$\mathcal{D}(\text{cars}) = \mathcal{D}(\text{sold}) \Vert
( \langle\text{sold}, \text{\sc Right}\rangle )$ (see~(3a)), 
$\mathcal{D}(\text{in}) = \mathcal{D}(\text{cars}) \Vert
( \langle\text{cars}, \text{\sc Nx-Right}\rangle )$ (according
to~(3b)).  \looseness=-1

\subsection{Tree Probability}

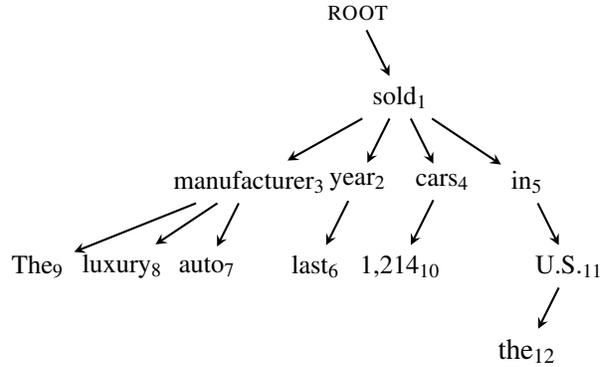
\begin{figure}[t]
\begin{tikzpicture}
[
scale=.75,
edge from parent/.style={draw,thick,-stealth},
]
\node {\small \sc root}
	child[missing]
	child {node {\small sold\textsubscript{1}}
		child {node[xshift=-9pt] {\small manufacturer\textsubscript{3}}
			child {node {\small The\textsubscript{9}}}
			child {node {\small luxury\textsubscript{8}}}
			child {node {\small auto\textsubscript{7}}}
			child [missing]
			child [missing]
			child [missing]
		}
		child {node {\small year\textsubscript{2}}
		    child {node {\small last\textsubscript{6}}}
		    child[missing]
		}
		child {node {\small cars\textsubscript{4}}
			child {node {\small 1,214\textsubscript{10}}}
			child[missing]
		}
		child {node {\small in\textsubscript{5}}
		    child[missing]
		    child {node {\small U.S.\textsubscript{11}}
		        child {node {the\textsubscript{12}}}
		        child[missing]
		    }
		}
	};
\end{tikzpicture}

\caption{Dependency tree of the sentence \textsl{The luxury auto manufacturer
    last year sold 1,214 cars in the U.S.} Subscripts
  indicate breadth-first traversal. \textsc{root} has only one
  dependent (i.e.,~\textsl{sold}) which we view as its first right
  dependent.}
\label{fig:deptree}
\end{figure}

The core problem in syntax-based language modeling is to estimate the
probability of sentence $S$ given its corresponding tree~$T$, $P(S | T)$.
We view the probability computation of a
dependency tree as a generation process. Specifically, we assume
dependency trees are constructed top-down, in a breadth-first
manner. Generation starts at the {\sc root} node. For each node at
each level, first its left dependents are generated from closest to
farthest and then the right dependents (again from closest to
farthest).  The same process is applied to the next node at the same
level or a node at the next level. Figure~\ref{fig:deptree} shows the
breadth-first traversal of a dependency tree.

Under the assumption that each word~$w$ in a dependency tree is
\emph{only} conditioned on its \emph{dependency path}, the probability
of a sentence~$S$ given its dependency tree $T$ is:
\begin{equation}
\label{eq:treeprob}
P(S | T) = \prod_{w \in \text{BFS}(T) \setminus \textsc{root}}^{} P(w|\mathcal{D}(w))
\end{equation}
where $\mathcal{D}(w)$ is the dependency path of $w$. Note that each
word~$w$ is visited according to its breadth-first search order
(\text{BFS}(T)) and the probability of {\sc root} is ignored since
every tree has one. The role of {\sc root} in a dependency tree is the
same as the begin of sentence token (BOS) in a sentence. When
computing $P(S|T)$ (or $P(S)$), the probability of \textsc{root} (or
BOS) is ignored (we assume it always exists), but is used to predict
other words. We explain in the next section how {\sc TreeLSTM} estimates~$P(w|\mathcal{D}(w))$. \looseness=-1


\subsection{Tree LSTMs}
\label{sec:treelstm}

\begin{figure}
\centering

\resizebox{0.48\textwidth}{!}{%
	
	\begin{tikzpicture}[scale=.75,->,>=stealth',thick,main node/.style={ellipse,fill=blue!20,draw,font=\sffamily\Large\bfseries,inner sep=0pt,minimum size=2.5mm}]
	\node[main node] (w0) at (0, 2) {$\mathbf{w_0}$};
	\node[main node] (w1) at (-2.5, 0) {$\mathbf{w_1}$};
	\node[main node] (w2) at (-6, 0) {$\mathbf{w_2}$};
	\node[main node] (w3) at (-9.5, 0) {$\mathbf{w_3}$};
	\node[main node] (w4) at (2.5, 0) {$\mathbf{w_4}$};
	\node[main node] (w5) at (6, 0) {$\mathbf{w_5}$};
	\node[main node] (w6) at (9.5, 0) {$\mathbf{w_6}$};
	
	\path (w0) edge[bend right=25,line width=1pt] (w1); 
	\path (w0) edge[bend right=15,line width=1pt] (w2); 
	\path (w0) edge[bend right=10,line width=1pt] (w3); 
	\path (w0) edge[bend left=25,line width=1pt] (w4); 
	\path (w0) edge[bend left=15,line width=1pt] (w5); 
	\path (w0) edge[bend left=10,line width=1pt] (w6);
	
	\draw[line width=2.5pt] (0, -1) -- node[left = 5pt, yshift=-2pt] {\bf Generated by four LSTMs} node[right = 5pt] {\bf with tied $\mathbf{W}_e$ and tied $\mathbf{W}_{ho}$} (0, -2.5);

	\end{tikzpicture}
	
} %

\vspace{0.25cm}

\resizebox{0.48\textwidth}{!}{%
	
	\centering
	\begin{tikzpicture}[scale=.75,->,>=stealth',thick,main node/.style={rectangle,rounded corners=3pt,fill=blue!10,draw,font=\sffamily\Large\bfseries,inner sep=0,minimum size=2.5mm,minimum width=4mm,minimum height=1.5cm,path picture={
			\draw[fill=blue!50!black] (0, -0.5) circle (1mm);
			\draw[fill=blue!50!black] (0, 0) circle (1mm);
			\draw[fill=blue!50!black] (0, 0.5) circle (1mm);
		}
	}]
	\node[main node] (w0) at (0, 2) {};
	\node[main node] (w1) at (-2.5, 0) {};
	\node[main node] (w2) at (-6, 0) {};
	
	\node (y0) at (0, -0.5) {\Large $\mathbf{w_0}$};
	
	\node (y1) at (-2.5, -2.5) {\Large $\mathbf{w_1}$};
	\node (y2) at (-6, -2.5) {\Large $\mathbf{w_2}$};
	\node[main node] (w3) at (-9.5, 0) {};
	\node (y3) at (-9.5, -2.5) {\Large $\mathbf{w_3}$};
	
	\node (x1) at (-2.5, 2.5) {\Large $\mathbf{w_0}$};
	\node (x2) at (-6, 2.5) {\Large $\mathbf{w_1}$};
	\node (x3) at (-9.5, 2.5) {\Large $\mathbf{w_2}$};
	
	\node[main node] (w4) at (2.5, 0) {};
	\node[main node] (w5) at (6, 0) {};
	\node[main node] (w6) at (9.5, 0) {};
	
	\node (y4) at (2.5, -2.5) {\Large $\mathbf{w_4}$};
	\node (y5) at (6, -2.5) {\Large $\mathbf{w_5}$};
	\node[main node] (w3) at (-9.5, 0) {};
	\node (y6) at (9.5, -2.5) {\Large $\mathbf{w_6}$};
	
	\node (x4) at (2.5, 2.5) {\Large $\mathbf{w_0}$};
	\node (x5) at (6, 2.5) {\Large $\mathbf{w_4}$};
	\node (x6) at (9.5, 2.5) {\Large $\mathbf{w_5}$};
	
	\draw[line width=1.5pt,blue] (w0) -- node[left = 5pt, above = 25pt, rotate=-50] {\textbf{ \textsc{Gen-L} }} (w1);
	\draw[line width=1.5pt] (w0) -- (y0);
	
	\draw[line width=1.5pt,red] (w1) -- node[above = 20pt] {\textbf{ \textsc{Gen-Nx-L} }} (w2);
	\draw[line width=1.5pt,blue] (x1) -- (w1);
	\draw[line width=1.5pt] (w1) -- (y1);
	
	\draw[line width=1.5pt,red] (w2) -- node[above = 20pt] {\textbf{\textsc{ Gen-Nx-L } }} (w3);
	\draw[line width=1.5pt,red] (x2) -- (w2);
	\draw[line width=1.5pt] (w2) -- (y2);
	\draw[line width=1.5pt,red] (x3) -- (w3);
	\draw[line width=1.5pt] (w3) -- (y3);
	
	\draw[line width=1.5pt,blue!70!black] (w0) -- node[right = 5pt, above = 25pt, rotate=50] {\textbf{ \textsc{Gen-R} }} (w4);
	\draw[line width=1.5pt,red!70!black] (w4) -- node[above = 20pt] {\textbf{\textsc{ Gen-Nx-R } }} (w5);
	\draw[line width=1.5pt,blue!70!black] (x4) -- (w4);
	\draw[line width=1.5pt] (w4) -- (y4);
	
	\draw[line width=1.5pt,red!70!black] (w5) -- node[above = 20pt] {\textbf{ \textsc{Gen-Nx-R} }} (w6);
	\draw[line width=1.5pt,red!70!black] (x5) -- (w5);
	\draw[line width=1.5pt] (w5) -- (y5);
	\draw[line width=1.5pt,red!70!black] (x6) -- (w6);
	\draw[line width=1.5pt] (w6) -- (y6);
	
	\end{tikzpicture}
	
}%
\vspace{-2mm}
\caption{Generation process of left ($w_1, w_2, w_3$) and right ($w_4,
  w_5, w_6$) dependents of tree node~$w_o$ (top) using four LSTMs
  ({\sc Gen-L}, {\sc Gen-R}, {\sc Gen-Nx-L} and {\sc Gen-Nx-R}).  The
  model can handle an arbitrary number of dependents due to {\sc
    Gen-Nx-L} and {\sc Gen-Nx-R}.}
\label{fig:treelstm}
\end{figure}
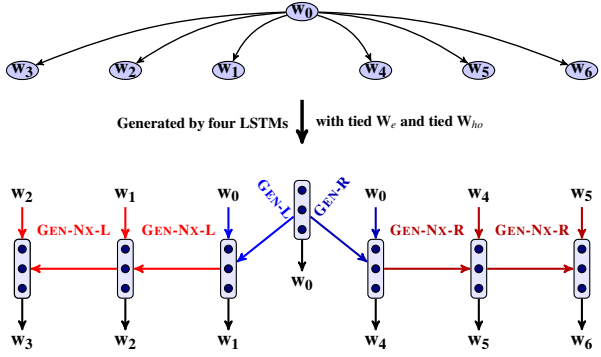

A dependency path $\mathcal{D}(w)$ is subtree which we denote as a
sequence of $\langle$\emph{word}, \emph{edge-type}$\rangle$
tuples. Our innovation is to learn the representation of
$\mathcal{D}(w)$ using four LSTMs.
The four LSTMs ({\sc Gen-L}, {\sc Gen-R}, {\sc Gen-Nx-L} and {\sc
  Gen-Nx-R}) are used to represent the four types of edges ({\sc
  Left}, {\sc Right}, {\sc Nx-Left} and \mbox{{\sc Nx-Right}})
introduced earlier. {\sc Gen}, {\sc Nx}, {\sc L} and {\sc R}~are
shorthands for {\sc Generate, Next, Left} and {\sc Right}. At each
time step, an LSTM is chosen according to an {edge-type}; then the
LSTM takes a word as input and predicts/generates its dependent or
sibling. This process can be also viewed as adding an edge and a node
to a tree. Specifically, LSTMs {\sc Gen-L} and {\sc Gen-R} are used to
generate the first left and right dependent of a node ($w_1$ and $w_4$
in Figure~\ref{fig:treelstm}). So, these two LSTMs are responsible for
going deeper in a tree. While {\sc Gen-Nx-L} and \mbox{{\sc Gen-Nx-R}}
generate the remaining left/right dependents and therefore go wider in
a tree. As shown in Figure~\ref{fig:treelstm}, $w_2$ and $w_3$ are
generated by \mbox{{\sc Gen-Nx-L}}, whereas $w_5$ and $w_6$ are
generated by \mbox{{\sc Gen-Nx-R}}. Note that the model can handle any
number of left or right dependents by applying \mbox{\sc Gen-Nx-L} or
\mbox{\sc Gen-Nx-R} multiple times.

We assume time steps correspond to the steps taken by the
breadth-first traversal of the dependency tree and the sentence has
length~$n$.  At time step~$t$ ($1 \leq t \leq n$), let $\langle
w_{t'}, z_t \rangle$ denote the last tuple
in~$\mathcal{D}(w_t)$. Subscripts~$t$ and $t'$ denote the
breadth-first search order of~$w_t$ and~$w_{t'}$, respectively. $z_t
\in \{ \text{\sc Left}, \text{\sc Right}, \text{\sc Nx-Left},
\text{\sc Nx-Right} \}$ is the edge type (see the definitions in
Section~\ref{sec:dpath}). Let $\mathbf{W}_e \in \mathbb{R}^{s \times
  |V|}$ denote the word embedding matrix and $\mathbf{W}_{ho} \in
\mathbb{R}^{|V|\times d}$ the output matrix of our model, where
$|V|$~is the vocabulary size, $s$~the word embedding size and $d$~the
hidden unit size. We use tied~$\mathbf{W}_e$ and tied
~$\mathbf{W}_{ho}$ for the four LSTMs to reduce the number of
parameters in our model. 
The four LSTMs also share their hidden states. Let~$\mathbf{H} \in
\mathbb{R}^{d \times (n+1)}$ denote the \emph{shared} hidden states of
all time steps and~$e(w_t)$ the one-hot vector of~$w_t$. Then,
$\mathbf{H}[:, t]$ represents~$\mathcal{D}(w_t)$ at time step~$t$, and
the computation\footnote{We ignore all bias terms for notational
  simplicity.} is:
\begin{subequations}
	\label{eq:treelstm}
	\begin{align}
		\mathbf{x}_t &= \mathbf{W}_e \cdot e(w_{t'}) \\
		\mathbf{h}_t &= \text{LSTM}^{z_t}(\mathbf{x}_t, \mathbf{H}[:,t']) \\
		\mathbf{H}{[:, t]} &= \mathbf{h}_t \\
		\mathbf{y}_t &= \mathbf{W}_{ho} \cdot \mathbf{h}_t
	\end{align}
\end{subequations}
where the initial hidden state $\mathbf{H}[:, 0]$ is initialized to a
vector of small values such as~0.01.  According to
Equation~(\ref{eq:treelstm}b), the model selects an LSTM based on edge
type~$z_t$. We describe the details of~$\text{LSTM}^{z_t}$ in the next
paragraph.  The probability of~$w_t$ given its dependency
path~$\mathcal{D}(w_t)$ is estimated by a \emph{softmax} function:
\begin{equation}
\label{eq:softmax}
P(w_t|\mathcal{D}(w_t)) = \frac{\exp(\mathbf{y}_{t,w_t})}{\sum_{k'=1}^{|V|} \exp(\mathbf{y}_{t,k'})}
\end{equation}
We must point out that although we use four jointly trained LSTMs to
encode the hidden states, the training and inference complexity of our
model is no different from a regular LSTM, since at each time step only
one LSTM is working.

We implement $\text{LSTM}^{z}$ in Equation~(\ref{eq:treelstm}b) using
a deep LSTM (to simplify notation, from now on we write~$z$ instead
of~$z_t$).  The inputs at time step~$t$ are $\mathbf{x}_t$ and
$\mathbf{h}_{t'}$ (the hidden state of an earlier time step~$t'$) and
the output is $\mathbf{h}_t$ (the hidden state of current time step).
Let~$L$ denote the layer number of $\text{LSTM}^{z}$ and
$\hat{\mathbf{h}}_t^{{l}}$ the internal hidden state of~the~$l$-th
layer of the $\text{LSTM}^{z}$ at time step $t$, where $\mathbf{x}_t$
is $\hat{\mathbf{h}}_t^{0}$ and $\mathbf{h}_{t'}$ is
$\hat{\mathbf{h}}_{t'}^L$.  The LSTM architecture introduces multiplicative gates and memory
cells $\hat{\mathbf{c}}^l_t$ (at $l$-th layer) in order to address the
\emph{vanishing gradient} problem which makes it difficult for the
standard RNN model to learn long-distance correlations in a
sequence. Here,~$\hat{\mathbf{c}}^l_t$ is a linear combination of the
current input signal $\mathbf{u}_t$ and an earlier memory
cell~$\hat{\mathbf{c}}^l_{t'}$. How much input information
$\mathbf{u}_t$ will flow into $\hat{\mathbf{c}}^l_t$ is controlled by
input gate $\mathbf{i}_t$ and how much of the earlier memory cell
$\hat{\mathbf{c}}^l_{t'}$ will be forgotten is controlled by forget
gate~$\mathbf{f}_t$.  This process is computed as follows:
\begin{subequations}
	\label{eq:lstm}
	\begin{align}
	\mathbf{u}_t &= \tanh( \mathbf{W}^{z,l}_{ux} \cdot \hat{\mathbf{h}}_t^{l-1} + \mathbf{W}^{z,l}_{uh} \cdot \hat{\mathbf{h}}_{t'}^l ) \\
	\mathbf{i}_t &= \sigma( \mathbf{W}^{z,l}_{ix} \cdot \hat{\mathbf{h}}_t^{l-1} + \mathbf{W}^{z,l}_{ih} \cdot \hat{\mathbf{h}}_{t'}^l ) \\
	\mathbf{f}_t &= \sigma( \mathbf{W}^{z,l}_{fx} \cdot \hat{\mathbf{h}}_t^{l-1} + \mathbf{W}^{z,l}_{fh} \cdot \hat{\mathbf{h}}_{t'}^l ) \\
	\hat{\mathbf{c}}^l_t &= \mathbf{f}_t \odot \hat{\mathbf{c}}^l_{t'} + \mathbf{i}_t \odot \mathbf{u}_t
	\end{align}
\end{subequations}
where $\mathbf{W}^{z,l}_{ux} \in \mathbb{R}^{d \times d}$
($\mathbf{W}^{z,l}_{ux} \in \mathbb{R}^{d \times s}$ when $l=1$) and
$\mathbf{W}^{z,l}_{uh} \in \mathbb{R}^{d \times d}$ are weight
matrices for $\mathbf{u}_t$, $\mathbf{W}^{z,l}_{ix}$ and
$\mathbf{W}^{z,l}_{ih}$ are weight matrices for $\mathbf{i}_t $ and
$\mathbf{W}^{z,l}_{fx}$, and $\mathbf{W}^{z,l}_{fh}$ are weight
matrices for $\mathbf{f}_t$. $\sigma$ is a sigmoid function and
$\odot$ the element-wise product. 

Output gate $\mathbf{o}_t$ controls how much information of the
cell~$\hat{\mathbf{c}}^l_t$ can be seen by other modules:
\begin{subequations}
	\label{eq:lstm}
	\begin{align}
		\mathbf{o}_t &= \sigma( \mathbf{W}^{z,l}_{ox} \cdot \hat{\mathbf{h}}_t^{l-1} + \mathbf{W}^{z,l}_{oh} \cdot \hat{\mathbf{h}}_{t'}^l ) \\
		\hat{\mathbf{h}}^l_t &= \mathbf{o}_t \odot \tanh(\hat{\mathbf{c}}^l_t)
	\end{align}
\end{subequations}
Application of the above process to all layers~$L$, will
yield~$\hat{\mathbf{h}}^L_t$, which is~$\mathbf{h}_t$.  Note that in
implementation, all $\hat{\mathbf{c}}^l_t$ and $\hat{\mathbf{h}}^l_t$
($1 \leq l \leq L$) at time step~$t$ are stored, although we only care
about $\hat{\mathbf{h}}^L_t$ ($\mathbf{h}_t$).

\subsection{Left Dependent Tree LSTMs}

\begin{figure}
\centering

\resizebox{0.48\textwidth}{!}{%

	\begin{tikzpicture}[scale=.75,->,>=stealth',thick,main node/.style={rectangle,rounded corners=3pt,fill=blue!10,draw,font=\sffamily\Large\bfseries,inner sep=0,minimum size=2.5mm,minimum width=4mm,minimum height=1.5cm,path picture={
			\draw[fill=blue!50!black] (0, -0.5) circle (1mm);
			\draw[fill=blue!50!black] (0, 0) circle (1mm);
			\draw[fill=blue!50!black] (0, 0.5) circle (1mm);
		}
	}]
	\node[main node] (w0) at (0, 2) {};
	\node[main node] (w1) at (-2.5, 0) {};
	\node[main node] (w2) at (-6, 0) {};
	\node[main node] (w3) at (-9.5, 0) {};
	
	\node (y0) at (0, -0.5) {\Large $\mathbf{w_0}$};

	\node (x1) at (-2.5, 2.5) {\Large $\mathbf{w_0}$};
	\node (x2) at (-6, 2.5) {\Large $\mathbf{w_1}$};
	\node (x3) at (-9.5, 2.5) {\Large $\mathbf{w_2}$};
	
	\node[main node] (w4) at (2.5, 0) {};
	\node[main node] (w5) at (6, 0) {};
	\node[main node] (w6) at (9.5, 0) {};
	
	\node (y4) at (2.5, -2.5) {\Large $\mathbf{w_4}$};
	\node (y5) at (6, -2.5) {\Large $\mathbf{w_5}$};
	\node[main node] (w3) at (-9.5, 0) {};
	\node (y6) at (9.5, -2.5) {\Large $\mathbf{w_6}$};
	
	\node (x4) at (2.5, 2.5) {\Large $\mathbf{w_0}$};
	\node (x5) at (6, 2.5) {\Large $\mathbf{w_4}$};
	\node (x6) at (9.5, 2.5) {\Large $\mathbf{w_5}$};
	
	\draw[line width=1.5pt,blue] (w0) -- node[left = 5pt, above = 25pt, rotate=-50] {\textbf{\textsc{Gen-L}}} (w1);
	\draw[line width=1.5pt] (w0) -- (y0);
	
	\draw[line width=1.5pt,red] (w1) -- node[above = 20pt] {\textbf{ \textsc{Gen-Nx-L} }} (w2);
	\draw[line width=1.5pt,blue] (x1) -- (w1);
	
	\draw[line width=1.5pt,red] (w2) -- node[above = 20pt] {\textbf{ \textsc{Gen-Nx-L} }} (w3);
	\draw[line width=1.5pt,red] (x2) -- (w2);
	\draw[line width=1.5pt,red] (x3) -- (w3);
	
	\draw[line width=1.5pt,blue!70!black] (w0) -- node[right = 5pt, above = 25pt, rotate=50] {\textbf{ \textsc{Gen-R} }} (w4);
	\draw[line width=1.5pt,red!70!black] (w4) -- node[above = 20pt] {\textbf{ \textsc{Gen-Nx-R} }} (w5);
	\draw[line width=1.5pt,blue!70!black] (x4) -- (w4);
	\draw[line width=1.5pt] (w4) -- (y4);
	
	\draw[line width=1.5pt,red!70!black] (w5) -- node[above = 20pt] {\textbf{ \textsc{Gen-Nx-R} }} (w6);
	\draw[line width=1.5pt,red!70!black] (x5) -- (w5);
	\draw[line width=1.5pt] (w5) -- (y5);
	\draw[line width=1.5pt,red!70!black] (x6) -- (w6);
	\draw[line width=1.5pt] (w6) -- (y6);

	\node[main node] (w1b) at (-2.5, -2.5) {};
	\node[main node] (w2b) at (-6, -2.5) {};
	\node[main node] (w3b) at (-9.5, -2.5) {};
	\node (wb) at (-11, -2.5) {};
	
	\node (y1) at (-2.5, -5) {\Large $\mathbf{w_1}$};
	\node (y2) at (-6, -5) {\Large $\mathbf{w_2}$};
	\node (y3) at (-9.5, -5) {\Large $\mathbf{w_3}$};
	
	\draw[line width=1.5pt,green!80!black] (wb) -- (w3b);
	\draw[line width=1.5pt,green!80!black] (w3b) -- node[above = 15pt] {\textbf{ \textsc{Ld} }} (w2b);
	\draw[line width=1.5pt,green!80!black] (w2b) -- node[above = 15pt] {\textbf{ \textsc{Ld} }} (w1b);
	\draw[line width=1.5pt,blue!70!black] (w1b) -- (w4);
	
	\draw[line width=1.5pt,green!80!black] (y1) -- (w1b);
	\draw[line width=1.5pt,green!80!black] (y2) -- (w2b);
	\draw[line width=1.5pt,green!80!black] (y3) -- (w3b);
	
	\path (w1) edge[bend right=25,line width=1.5pt] (y1); 
	\path (w2) edge[bend right=25,line width=1.5pt] (y2);
	\path (w3) edge[bend right=25,line width=1.5pt] (y3); 
	
	\end{tikzpicture}
	
}%
\vspace{-2mm}
\caption{Generation of left and right dependents of node $w_0$
  according to {\sc LdTreeLSTM}.}
\label{fig:bitreelstm}
\end{figure}
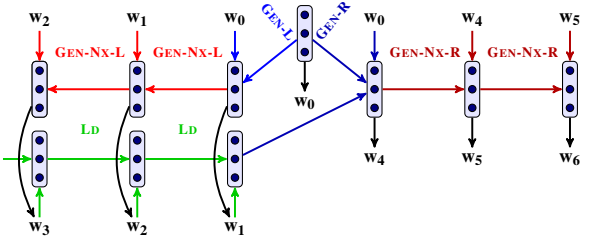

{\sc TreeLSTM} computes $P(w|\mathcal{D}(w))$ based on the dependency
path $\mathcal{D}(w)$, which ignores the interaction between left and
right dependents on the same level. In many cases, {\sc TreeLSTM} will
use a verb to predict its object directly without knowing its subject.
For example, in Figure~\ref{fig:deptree}, {\sc TreeLSTM} uses
$\langle${\sc root}, {\sc Right}$\rangle$ and \mbox{$\langle$
  \emph{sold}, {\sc Right} $\rangle$} to predict \emph{cars}. This
information is unfortunately not specific to \textsl{cars} (many
things can be sold, e.g.,~\textsl{chocolates},
\textsl{candy}). Considering \textsl{manufacturer}, the left dependent
of \textsl{sold} would help predict \textsl{cars} more accurately.

In order to jointly take left and right dependents into account, we
employ yet another LSTM, which goes from the furthest left dependent
to the closest left dependent ({\sc Ld} is a shorthand for left
dependent). As shown in Figure~\ref{fig:bitreelstm}, {\sc Ld} LSTM
learns the representation of all left dependents of a node $w_0$; this
representation is then used to predict the first right dependent of
the same node.  Non-first right dependents can also leverage the
representation of left dependents, since this information is injected
into the hidden state of the first right dependent and can percolate
all the way. Note that in order to retain the generation capability of
our model (Section~\ref{sec:treegen}), we only allow right
dependents to leverage left dependents (they are generated before right dependents).

The computation of the {\sc LdTreeLSTM} is almost the same as in {\sc
  TreeLSTM} except when \mbox{$z_t = \text{\sc Gen-R}$}.  In this
case, let $\mathbf{v}_t$ be the corresponding left dependent sequence
with length \mbox{$K$ ($\mathbf{v}_t=(w_3,w_2,w_1)$} in
Figure~\ref{fig:bitreelstm}).  Then, the hidden state ($\mathbf{q}_k$)
of $\mathbf{v}_t$ at each time step $k$ is:
\begin{subequations}
	\label{eq:lch}
	\begin{align}
	\mathbf{m}_k &= \mathbf{W}_e \cdot e(\mathbf{v}_{t,k}) \\
	\mathbf{q}_k &= \text{LSTM}^{\text{\sc Ld}}(\mathbf{m}_k, \mathbf{q}_{k-1})
	\end{align}
\end{subequations}
where~$\mathbf{q}_K$ is the representation for all left dependents. Then, the
computation of the current hidden state becomes (see
Equation~\eqref{eq:treelstm} for the original computation):
\begin{subequations}
	\label{eq:bitreelstm}
	\begin{align}
	\mathbf{r}_t &= \begin{bmatrix}
		\mathbf{W}_e \cdot e(w_{t'}) \\
		\mathbf{q}_K
	\end{bmatrix} \\
	\mathbf{h}_t &= \text{LSTM}^{\text{\sc Gen-R}}(\mathbf{r}_t, \mathbf{H}[:,t'])
	\end{align}
\end{subequations}
where $\mathbf{q}_K$ serves as additional input for
$\text{LSTM}^{\text{\sc Gen-R}}$.  All other computational details are
the same as in {\textsc TreeLSTM} (see Section~\ref{sec:treelstm}).

\subsection{Model Training}

On small scale datasets we employ Negative Log-likelihood (NLL) as our
training objective for both {\sc TreeLSTM} and {\sc LdTreeLSTM}:
\begin{equation}
\label{eq:nllobj}
\mathcal{L}^{\text{NLL}}(\theta) = -\frac{1}{|\mathcal{S}|} \sum_{S \in \mathcal{S}} \log P(S | T)
\end{equation}
where $S$ is a sentence in the training set $\mathcal{S}$, $T$ is the
dependency tree of $S$ and $P(S | T)$~is defined as in
Equation~\eqref{eq:treeprob}.

On large scale datasets (e.g.,~with vocabulary size of~65K), computing
the output layer activations and the \emph{softmax} function with NLL would
become prohibitively expensive.  Instead, we employ Noise Contrastive
Estimation (NCE; \newcite{gutmann2012noise}, \newcite{MnihTeh2012})
which treats the normalization term $\hat{Z}$ in
$\hat{P}(w|\mathcal{D}(w_t)) = \frac{\exp(\mathbf{W}_{ho}[w,:] \cdot
  \mathbf{h}_t)}{\hat{Z}}$ as constant.  The intuition behind NCE is
to discriminate between samples from a data distribution
$\hat{P}(w|\mathcal{D}(w_t))$ and a known noise distribution $P_n(w)$
via binary logistic regression.  Assuming that noise words are
$k$~times more frequent than real words in the training set
\cite{MnihTeh2012}, then the probability of a word~$w$ being from our
model $P_d(w, \mathcal{D}(w_t))$ is
$\frac{\hat{P}(w|\mathcal{D}(w_t))}{ \hat{P}(w|\mathcal{D}(w_t)) + k
  P_n(w) }$.
We apply NCE to large vocabulary models with the following training
objective:
\begin{equation*}
\begin{split}
\mathcal{L}^{\text{NCE}}(\theta) = &
-\frac{1}{|\mathcal{S}|} \sum_{T \in \mathcal{S}} \sum_{t=1}^{|T|}  \bigg( \log P_d(w_t, \mathcal{D}(w_t)) \biggr. \\
& \biggl. + \sum_{j=1}^k \log [ 1 - P_d(\tilde{w}_{t,j}, \mathcal{D}(w_t)) ] \biggr) 
\end{split}
\end{equation*}
where $\tilde{w}_{t,j}$ is a word sampled from the noise
distribution~$P_n(w)$. We use smoothed unigram frequencies
(exponentiating by~0.75) as the noise distribution~$P_n(w)$
\cite{Mikolov:ea:2013a}. We initialize $\ln \hat{Z} = 9$ as suggested
in \newcite{chen2015recurrent}, but instead of keeping it fixed we
also learn $\hat{Z}$ during training \cite{vaswani2013decoding}.  We
set $k=20$.

%

\section{Experiments}

We assess the performance of our model on two tasks: the Microsoft
Research (MSR) sentence completion challenge
\cite{zweig2012challenge}, and dependency parsing reranking.  We also
demonstrate the tree generation capability of our models.  In the
following, we first present details on model training and then present
our results. We implemented our models using the Torch library
\cite{collobert2011torch7} and our code is available at
\url{https://github.com/XingxingZhang/td-treelstm}.

\subsection{Training Details}
We trained our model with back propagation through time
\cite{rumelhart1988learning} on an Nvidia GPU Card with a mini-batch
size of~64.  The objective (NLL or NCE) was minimized by stochastic
gradient descent. Model parameters were uniformly initialized in
$[-0.1, 0.1]$. We used the NCE objective on the MSR sentence
completion task (due to the large size of this dataset) and the NLL
objective on dependency parsing reranking. We used an initial learning
rate of~1.0 for all experiments and when there was no significant
improvement in log-likelihood on the validation set, the learning rate
was divided by~$2$ per epoch until convergence
\cite{mikolov2010recurrent}.  To alleviate the exploding gradients
problem, we rescaled the gradient~$g$ when the gradient norm~$||g|| >
5$ and set $g = \frac{5g}{|| g ||}$
\cite{pascanu2013difficulty,sutskever2014sequence}.  Dropout
\cite{srivastava2014dropout} was applied to the 2-layer {\sc TreeLSTM}
and {\sc LdTreeLSTM} models.
The word embedding size was set to~$s = d
/ 2$ where $d$ is the hidden unit size.

\subsection{Microsoft Sentence Completion Challenge}
\label{sec:msr}

The task in the MSR Sentence Completion Challenge
\cite{zweig2012challenge} is to select the correct missing word for
1,040 \mbox{SAT-style} test sentences when presented with five
candidate completions. The training set contains 522 novels from the
Project Gutenberg which we preprocessed as follows.  After removing
headers and footers from the files, we tokenized and parsed the
dataset into dependency trees with the Stanford Core NLP toolkit
\cite{manning2014stanford}. The resulting training set contained 49M
words. We converted all words to lower case and replaced those
occurring five times or less with UNK. The resulting vocabulary size
was~65,346 words. We randomly sampled 4,000 sentences from the
training set as our validation set.


The literature describes two main approaches to the sentence
completion task based on word vectors and language models. In
vector-based approaches, all words in the sentence and the five
candidate words are represented by a vector; the candidate which has
the highest average similarity with the sentence words is selected as
the answer.  For language model-based methods, the LM computes the
probability of a test sentence with each of the five candidate words,
and picks the candidate completion which gives the highest
probability.  Our model belongs to this class of models.

\begin{table}[t!]
\centering
\small
\begin{tabular}{ l |  c  c | c }
\hline
{\bf Model} & $d$ & $| \theta |$ & {\bf Accuracy} \\
\hline
\hline
\multicolumn{4}{l}{Word Vector based Models} \\
\hline
LSA & --- & --- & 49.0\\
Skip-gram & 640 & 102M & 48.0 \\
{\sc ivLBL} & 600 & 96.0M & 55.5 \\
\hline
\hline
\multicolumn{4}{l}{Language Models} \\
\hline
KN5 & --- & --- & 40.0 \\
UDepNgram & --- & --- & 48.3 \\
LDepNgram & ---& --- & 50.0 \\ 
RNN & 300 & 48.1M & 45.0 \\
RNNME & 300 & 1120M & 49.3 \\
depRNN+3gram  & 100 & 1014M & 53.5 \\
ldepRNN+4gram & 200 & 1029M & 50.7 \\
LBL &  300 & 48.0M & 54.7 \\
\hline
LSTM & 300 & 29.9M & 55.00 \\
LSTM & 400 & 40.2M & 57.02 \\
LSTM & 450 & 45.3M & 55.96 \\
Bidirectional LSTM & 200 & 33.2M & 48.46 \\
Bidirectional LSTM & 300 & 50.1M & 49.90 \\
Bidirectional LSTM & 400 & 67.3M & 48.65 \\
\hline \hline
\multicolumn{4}{l}{Model Combinations} \\
\hline
RNNMEs & --- & --- & 55.4 \\
Skip-gram + RNNMEs & --- & --- & 58.9 \\
\hline
\hline
\multicolumn{4}{l}{Our Models} \\
\hline

{\sc TreeLSTM} & 300 & 31.6M & 55.29 \\
{\sc LdTreeLSTM} & 300 & 32.5M & {57.79} \\

{\sc TreeLSTM} & 400 & 43.1M & 56.73 \\
{\sc LdTreeLSTM} & 400 & 44.7M & {\bf 60.67} \\
\hline
\end{tabular}
\caption{Model accuracy on the MSR sentence completion task. The results
  of KN5, RNNME and RNNMEs are reported in
  \protect\newcite{mikolov2012thesis},  LSA and RNN in
  \protect\newcite{zweig2012computational}, UDepNgram and LDepNgram in
  \protect\newcite{gubbins-vlachos:2013:EMNLP},  depRNN+3gram and
  depRNN+4gram in \protect\newcite{mirowski-vlachos:2015}, LBL in
  \protect\newcite{MnihTeh2012}, Skip-gram and Skip-gram+RNNMEs in
  \protect\newcite{Mikolov:ea:2013b}, and  {\sc ivLBL} in
  \protect\newcite{NIPS2013_5165}; $d$ is the hidden size and 
  $|\theta|$~the number of parameters in a model.}
\label{tab:msr}
\end{table}

Table~\ref{tab:msr} presents a summary of our results together with
previoulsy published results. The best performing word vector model is
{\sc ivLBL} \cite{NIPS2013_5165} with an accuracy of~55.5, while the
best performing single language model is {\sc LBL} \cite{MnihTeh2012}
with an accuracy of~54.7. Both approaches are based on the
log-bilinear language model \cite{MnihHinton2007}. A combination of
several recurrent neural networks and the skip-gram model holds the
state of the art with an accuracy of~58.9 \cite{Mikolov:ea:2013a}. To
fairly compare with existing models, we restrict the layer size of our
models to 1. We observe that {\sc LdTreeLSTM} consistently outperforms
{\sc TreeLSTM}, which indicates the importance of modeling the
interaction between left and right dependents. In fact, {\sc
  LdTreeLSTM} ($d=400$) achieves a new state-of-the-art on this task,
despite being a single model. We also implement LSTM and bidirectional
LSTM language models.\footnote{LSTMs and BiLSTMs were also trained with
 NCE (\mbox{$s=d/2$}; hyperparameters were tuned on the
  development set).} An LSTM with $d=400$ outperforms its smaller
counterpart ($d=300$), however performance decreases with $d=450$. The
bidirectional LSTM is worse than the LSTM (see \newcite{MnihTeh2012}
for a similar observation). The best performing LSTM is worse than a
{\sc LdTreeLSTM} ($d=300$). The input and output embeddings
($\mathbf{W}_e$ and $\mathbf{W}_{ho}$) dominate the number of
parameters in all neural models except for RNNME, depRNN+3gram and
ldepRNN+4gram, which include a ME model that contains 1 billion sparse
n-gram features \cite{mikolov2012thesis,mirowski-vlachos:2015}. The
number of parameters in {\sc TreeLSTM} and {\sc LdTreeLSTM} is not
much larger compared to LSTM due to the tied $\mathbf{W}_e$
and~$\mathbf{W}_{ho}$ matrices.

\subsection{Dependency Parsing}
\label{sec:depparse}

\begin{table}[t]
\centering
\begin{tabular}{ l|cc|cc }
\hline
\multirow{2}{*}{\bf Parser}         & \multicolumn{2}{c|}{Development} & \multicolumn{2}{c}{Test} \\
  & UAS & LAS & UAS & LAS \\
\hline
\hline
MSTParser-2nd   & 92.20 & 88.78 & 91.63 & 88.44 \\
{\sc TreeLSTM} & 92.51 & 89.07 & 91.79 &  88.53 \\
{\sc TreeLSTM*} & 92.64 & 89.09 & 91.97 &  88.69 \\
{\sc LdTreeLSTM} & {\bf 92.66} & {\bf 89.14} & {\bf 91.99} & {\bf
  88.69} \\ \hline\hline
NN parser* & 92.00  & 89.70  & 91.80  & 89.60  \\
S-LSTM*    & {\bf 93.20}  & {\bf 90.90}  & {\bf 93.10}  & {\bf 90.90}  \\
\hline
\end{tabular}
\caption{Performance of {\sc TreeLSTM} and {\sc LdTreeLSTM} on
  reranking the top dependency trees produced by the 2nd order MSTParser
  \protect\cite{mcdonald2006online}. 
  Results for the NN and S-LSTM parsers are reported in
  \protect\newcite{chen-manning:2014:EMNLP2014} and
  \protect\newcite{dyer-EtAl:2015:ACL-IJCNLP}, respectively. *
  indicates that the model is initialized with pre-trained word vectors.}
\label{tab:depparse}
\end{table}

In this section we demonstrate that our model can be also used for
parse reranking. This is not possible for sequence-based language
models since they cannot estimate the probability of a tree. We use
our models to rerank the top~$K$ dependency trees produced by the
second order MSTParser
\cite{mcdonald2006online}.\footnote{http://www.seas.upenn.edu/~strctlrn/MSTParser}
We follow closely the experimental setup of
\newcite{chen-manning:2014:EMNLP2014} and
\newcite{dyer-EtAl:2015:ACL-IJCNLP}. Specifically, we trained {\sc
  TreeLSTM} and {\sc LdTreeLSTM} on Penn Treebank
sections~\mbox{2--21}. We used section~22 for development and
section~23 for testing. We adopted the Stanford basic dependency
representations \cite{de2006generating}; part-of-speech tags were
predicted with the Stanford Tagger \cite{toutanova2003feature}.  We
trained {\sc TreeLSTM} and {\sc LdTreeLSTM} as language models
(singletons were replaced with UNK) and did not use any POS tags,
dependency labels or composition features, whereas these features are used in
\newcite{chen-manning:2014:EMNLP2014} and
\newcite{dyer-EtAl:2015:ACL-IJCNLP}.  We tuned~$d$, the number of
layers, and~$K$ on the development set.

Table~\ref{tab:depparse} reports unlabeled attachment scores (UAS) and
labeled attachment scores (LAS) for the MSTParser, {\sc TreeLSTM}
($d=300$, 1 layer, $K=2$), and {\sc LdTreeLSTM} ($d=200$, 2 layers,
\mbox{$K=4$}).  We also include the performance of two neural
network-based dependency parsers;
\newcite{chen-manning:2014:EMNLP2014} use a neural network classifier
to predict the correct transition (NN parser);
\newcite{dyer-EtAl:2015:ACL-IJCNLP} also implement a transition-based
dependency parser using LSTMs to represent the contents of the stack and buffer in a continuous space.  As can be seen, both {\sc
  TreeLSTM} and {\sc LdTreeLSTM} outperform the baseline MSTParser,
with {\sc LdTreeLSTM} performing best.  We also initialized the word
embedding matrix $\mathbf{W}_e$ with pre-trained GLOVE vectors
\cite{pennington2014glove}. We obtained a slight improvement over {\sc
  TreeLSTM} ({\sc TreeLSTM*} in Table~\ref{tab:depparse}; $d=200$, 2
layer, $K=4$) but no improvement over {\sc LdTreeLSTM}. Finally,
notice that {\sc LdTreeLSTM} is slightly better than the NN parser in
terms of UAS but worse than the S-LSTM parser. In the future, we would
like to extend our model so that it takes labeled dependency
information into
account.

\subsection{Tree Generation}
\label{sec:treegen}

\begin{figure}
	\includegraphics[width=\columnwidth]{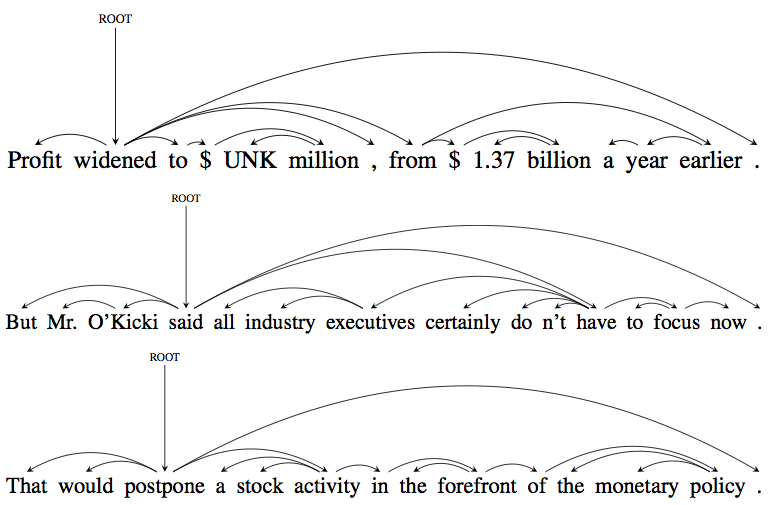}
	\caption{Generated dependency trees with {\sc LdTreeLSTM} trained on the PTB.}
	\label{fig:treesample}
\end{figure}

This section demonstrates how to use a trained {\sc LdTreeLSTM} to
generate tree samples. The generation starts at the {\sc root}
node. At each time step~$t$, for each node $w_t$, we add a new edge
and node to the tree. Unfortunately during generation, we do not know
which type of edge to add. We therefore use four binary classifiers
({\sc Add-Left}, {\sc Add-Right}, {\sc Add-Nx-Left} and {\sc
  Add-Nx-Right}) to predict whether we should add a {\sc Left}, {\sc
  Right}, {\sc Nx-Left} or {\sc Nx-Right} edge.\footnote{It is
  possible to get rid of the four classifiers by adding START/STOP
  symbols when generating left and right dependents as in
  \cite{eisner1996three}. We refrained from doing this for
  computational reasons.  For a sentence with~$N$ words, this approach
  will lead to~$2N$ additional START/STOP symbols (with one START and
  one STOP symbol for each word).  Consequently, the computational
  cost and memory consumption during training will be three times as
  much rendering our model less scalable.  } Then when a classifier
predicts true, we use the corresponding LSTM to generate a new node by
sampling from the predicted word distribution in
Equation~\eqref{eq:softmax}. The four classifiers take the previous
hidden state $\mathbf{H}[:,t']$ and the output embedding of the
current node $\mathbf{W}_{ho} \cdot e(w_{t})$ as
features.\footnote{The input embeddings have lower dimensions and
  therefore result in slightly worse classifiers.} Specifically, we
use a trained {\sc LdTreeLSTM} to go through the training corpus and
generate hidden states and embeddings as input features; the
corresponding class labels (true and false) are ``read off'' the
training dependency trees.  We use two-layer rectifier networks
\cite{glorot2011deep} as the four classifiers with a hidden size
of~300. We use the same {\sc LdTreeLSTM} model as in
Section~\ref{sec:depparse} to generate dependency trees. The
classifiers were trained using AdaGrad \cite{duchi2011adaptive} with a
learning rate of~0.01. The accuracies of {\sc Add-Left}, {\sc
  Add-Right}, {\sc Add-Nx-Left} and {\sc Add-Nx-Right} are 94.3\%,
92.6\%, 93.4\% and 96.0\%, respectively. Figure~\ref{fig:treesample}
shows examples of generated trees.

\section{Conclusions}
In this paper we developed {\sc TreeLSTM} (and {\sc LdTreeLSTM}), a
neural network model architecture, which is designed to predict tree
structures rather than linear sequences. Experimental results on the
MSR sentence completion task show that {\sc LdTreeLSTM} is superior to
sequential LSTMs. Dependency parsing reranking experiments highlight
our model's potential for dependency parsing.  Finally, the ability of
our model to generate dependency trees holds promise for text
generation applications such as sentence compression and
simplification \cite{filippova2015sentence}. Although our experiments
have focused exclusively on dependency trees, there is nothing
inherent in our formulation that disallows its application to other
types of tree structure such as constituent trees or even taxonomies.

\section*{Acknowledgments}
We would like to thank Adam Lopez, Frank Keller, Iain Murray, Li Dong,
Brian Roark, and the NAACL reviewers for their valuable feedback.
Xingxing Zhang gratefully acknowledges the financial support of the
China Scholarship Council (CSC).  Liang Lu is funded by the UK EPSRC
Programme Grant EP/I031022/1, Natural Speech Technology (NST).


\bibliography{naaclhlt2016}
\bibliographystyle{naaclhlt2016}

\end{document}